\documentclass[conference, 10pt]{IEEEtran}

\ifCLASSOPTIONcompsoc
  \usepackage[caption=false,font=normalsize,labelfont=sf,textfont=sf]{subfig}
\else
  \usepackage[caption=false,font=footnotesize]{subfig}
\fi
\usepackage{ifpdf}
\usepackage{fixltx2e}
\usepackage{stfloats}
\usepackage{url}
\usepackage{cite}
\usepackage{graphicx}
\usepackage{textcomp}
\def\BibTeX{{\rm B\kern-.05em{\sc i\kern-.025em b}\kern-.08em
    T\kern-.1667em\lower.7ex\hbox{E}\kern-.125emX}}
\usepackage{amsmath,amssymb,amsfonts}
\usepackage{booktabs} 
\usepackage{multirow}
\usepackage{xcolor}
\usepackage[super]{nth}
\usepackage{tikz}
\newcommand*\circled[1]{\tikz[baseline=(char.base)]{
            \node[shape=circle,draw,inner sep=0.2pt] (char) {#1};}}
\newcommand*\circledB[1]{\tikz[baseline=(char.base)]{
            \node[shape=circle,fill,inner sep=0.2pt] (char) {\textcolor{white}{#1}};}}

\usepackage{soul}
\usepackage{enumitem}
\usepackage{algorithm}
\usepackage[noend]{algorithmic}
\usepackage{array}

\usepackage{fancyhdr}
\fancypagestyle{firstpage}{
  \fancyhf{}
  \chead{To appear at the 2021 International Joint Conference on Neural Networks (IJCNN), July 2021, Virtual Event.}
  \cfoot{\thepage}
}

\usepackage[none]{hyphenat}
\begin{document}

\title{Q-SpiNN: A Framework for Quantizing \\ Spiking Neural Networks
\vspace{-0.4cm}}

\author{\IEEEauthorblockN{Rachmad Vidya Wicaksana Putra\textsuperscript{1}, Muhammad Shafique\textsuperscript{2}}
\IEEEauthorblockA{\textsuperscript{1}\textit{Technische Universit\"at Wien (TU Wien)}, Vienna, Austria \\
\textit{\textsuperscript{2}New York University Abu Dhabi (NYUAD)}, Abu Dhabi, United Arab Emirates \\
Email: rachmad.putra@tuwien.ac.at,
muhammad.shafique@nyu.edu}
\vspace{-1.1cm}
}

\maketitle
\thispagestyle{firstpage}

\begin{abstract}
A prominent technique for reducing the memory footprint of Spiking Neural Networks (SNNs) without decreasing the accuracy significantly is quantization. 
However, the state-of- the-art only focus on employing the weight quantization directly from a specific quantization scheme, i.e., either the post-training quantization (PTQ) or the in-training quantization (ITQ), and do not consider 
(1) quantizing other SNN parameters (e.g., neurons' membrane potential), 
(2) exploring different combinations of quantization approaches (i.e., quantization schemes, precision levels, and rounding schemes), and 
(3) selecting the SNN model with a good memory-accuracy trade-off at the end. 
Therefore, the memory saving offered by these state-of-the-art to meet the targeted accuracy is limited, thereby hindering processing SNNs on the resource-constrained systems (e.g., the IoT-Edge devices).
Towards this, we propose Q-SpiNN, a novel quantization framework for memory-efficient SNNs. 
The key mechanisms of the Q-SpiNN are: 
(1) employing quantization for different SNN parameters based on their significance to the accuracy, 
(2) exploring different combinations of quantization schemes, precision levels, and rounding schemes to find efficient  SNN model candidates, and 
(3) developing an algorithm that quantifies the benefit of the memory-accuracy trade-off obtained by the candidates, and selects the Pareto-optimal one. 
The experimental results show that, for the unsupervised network, the Q-SpiNN reduces the memory footprint by ca. 4x, while maintaining the accuracy within 1\% from the baseline on the MNIST dataset. 
For the supervised network, the Q-SpiNN reduces the memory by ca. 2x, while keeping the accuracy within 2\% from the baseline on the DVS-Gesture dataset.  
\end{abstract}

% \vspace{-0.3cm}
% \begin{IEEEkeywords}
% Spiking neural networks, quantization, model compression, memory saving, memory-accuracy trade-off. 
% % resource-constrained systems.
% \vspace{-0.2cm}
% \end{IEEEkeywords}


% \fi
\IEEEpeerreviewmaketitle

%%%%%%%%%%%%%%%%%%%%%%%%%%%%%%%%%%%%%%%%%%%%%%%%%%%%%%%%%%%%%%%%%%%%%%%%%%%%%%%%%%%%%%%%%%%%%%%%%%%%%%%%%%%%%%%%%%%%%%%%%%%%%%%%%%%%%%%%%%%%%%%%%%%%%%%%%%%%%%%%%%%%%%%%
% \vspace{-0.1cm}
\section{Introduction}
\label{Sec_Intro}
\vspace{-0.1cm}
 
\begin{figure}[t]
% \vspace{-0.3cm}
\centering
\includegraphics[width=\linewidth]{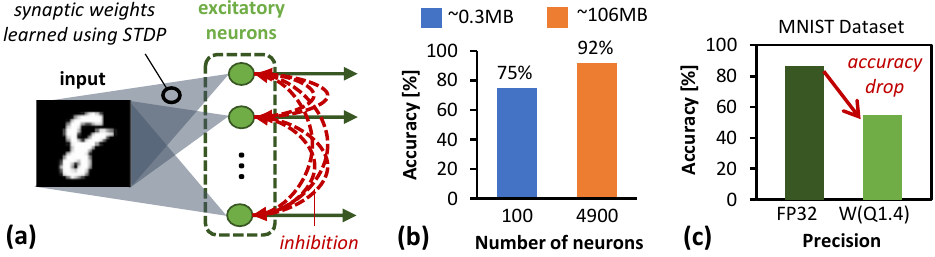}
\vspace{-0.7cm}
\caption{(a) A fully-connected SNN architecture with the spike-timing- dependent plasticity based unsupervised learning (U-SNN), and its detailed information will be discussed further in Section~\ref{Sec_Background_SNNs}. 
(b) Accuracy and memory footprint of the U-SNN with different number of neurons. 
(c) Accuracy of the 400-neurons U-SNN with different precision levels.} % over the MNIST test set.} 
% Here, \texttt{FP32} denotes 32-bit floating-point, while W(\texttt{Q1.4}) denotes the 6-bit fixed-point with 1-bit sign, 1-bit integer, 4-bit fractional for weights. }
\label{Fig_ObserveQuant}
\vspace{-0.6cm}
\end{figure}

SNN models have been proposed to solve various data analytic tasks, such as digit classification, object detection, and hand gesture recognition \cite{Ref_Pfeiffer_DLSNN_FNINS18, Ref_Tavanaei_DLSNN_Neunet18, Ref_Diehl_STDPmnist_FNCOM15,Ref_Hazan_SOMSNN_IJCNN18,Ref_Saunders_STDPpatch_IJCNN18,Ref_Saunders_LCSNN_NeuNet19,Ref_Hazan_LMSNN_AMAI19,Ref_Kaiser_DECOLLE_FNINS20, Ref_Massa_SNN_IJCNN20, Ref_Putra_SparkXD_arXiv21, Ref_Venceslai_NeuroAttack_IJCNN20, Ref_Putra_SpikeDyn_arXiv21}.
To achieve high accuracy, many large-sized SNN models have been developed, as they have shown better capability for learning the input features than the small ones. 
For instance, a large SNN model that occupies $>$100MB of memory with 32-bit floating-point format (\texttt{FP32}), achieves ca. 92\% accuracy on MNIST dataset \cite{Ref_Lecun_MNIST_IEEE98}.  
Meanwhile, a small model that occupies ca. 0.3MB of memory with \texttt{FP32}, achieves ca. 75\% accuracy, as shown in Figs.~\ref{Fig_ObserveQuant}(a)-(b). 
Consequently, the state-of-the-art SNN models typically have large number of parameters that need to be considered in both the training and the inference phases.  
Therefore, they incur large memory footprint, which hinder their applicability for the resource-constrained systems, such as the IoT-Edge devices.

To address these issues, prior works have proposed different methodologies, 
such as (1) \textit{reduction of SNN operations} via stochastic neuron operations \cite{Ref_Sen_ApproxSNN_DATE17}, neuron elimination \cite{Ref_Putra_FSpiNN_TCAD20}, and weight pruning \cite{Ref_Rathi_PruneQuantizeSNN_TCAD18}; and (2) \textit{quantization} \cite{Ref_Rathi_PruneQuantizeSNN_TCAD18, Ref_Sorbaro_OptimSNN_FNINS20,Ref_Zou_MedianQuant_ISCAS20}.
Among these techniques, quantization is a prominent one that incurs relatively low overhead, since it only needs to reduce the data precision. 
Besides memory saving, the reduced precision also leads to other advantages, e.g., faster computation and lower power/energy consumption. 
\textit{However, reducing the precision of SNN parameters leads to accuracy degradation if it is not performed carefully due to the information loss}, as shown in Fig.~\ref{Fig_ObserveQuant}(c).
The results show that a network with W(\texttt{Q1.4}) or 6-bit of fixed-point weights, suffers from accuracy drop, compared to the 32-bit floating-point (\texttt{FP32}).  
Here, the \texttt{Q1.4} format denotes 1 sign bit, 1 integer bit, and 4 fractional bits\footnote{\textbf{Note:} In this paper, the fixed-point format is represented as \texttt{Qi.f}, with 1 sign bit, \texttt{i} integer bits, and \texttt{f} fractional bits \cite{Ref_Granas_FixedPointTheory}. The value of \texttt{i} for each parameter depends on the range of its integer values. The detailed discussion of the fixed-point format is presented in Section~\ref{Sec_Background_FixedPoint}.}. 

\textbf{Targeted Research Problem:} 
\textit{If and how can we employ the quantization on SNNs to maximize the memory saving, while maintaining the accuracy.}
An efficient solution to this problem will improve the applicability of the SNN systems on resource-constrained devices. 

%%%%%%%%%%%%%%%%%
%%%%%%%%%%%%%%%%%
\subsection{State-of-the-art and Their Limitations}
\label{Sec_SOTA}
\vspace{-0.1cm}

The state-of-the-art have employed quantization to reduce the precision of the weights by directly using a specific quantization scheme, i.e., either the post-training quantization (PTQ) or the in-training quantization (ITQ) \cite{Ref_Rathi_PruneQuantizeSNN_TCAD18,Ref_Sorbaro_OptimSNN_FNINS20,Ref_Zou_MedianQuant_ISCAS20}. 
% and showed that a relatively high accuracy can still be achieved. 
However, they have several drawbacks as they \textit{do not} consider: 
% discussed below.
% 
% \textbf{Limitations:} 
% These state-of-the-art works \textit{do not} consider: 
\begin{itemize}[leftmargin=*]
    \item quantization for other SNN parameters (e.g., the neurons' membrane potential) that occupy a considerable amount of memory during the SNN processing \cite{Ref_Sen_ApproxSNN_DATE17}\cite{Ref_Roy_PEASE_ISLPED17}, 
    \item exploring different combinations of \textit{quantization approaches} (i.e., quantization schemes, precision levels, and rounding schemes) to find the SNN model that fulfills the targeted accuracy and achieves high memory saving. 
\end{itemize}
Therefore, \textit{the memory saving offered by these state-of-the-art to meet the targeted accuracy are limited, thereby hindering the deployment of SNNs on the resource-constrained devices.}
To highlight the targeted problem and the limitations of the state-of-the-art, we perform an experimental case study, as discussed below.

%%%%%%%%%%%%%%%%%
%%%%%%%%%%%%%%%%%
\vspace{-0.1cm}
\subsection{Motivational Case Study and Key Challenges}
\label{Sec_CaseStudy}
\vspace{-0.1cm}

\begin{figure}[t]
% \vspace{-0.3cm}
\centering
\includegraphics[width=0.9\linewidth]{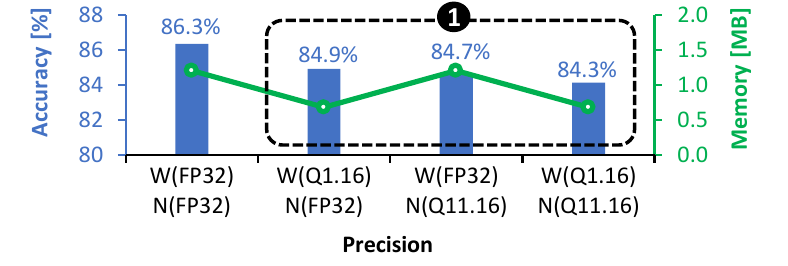}
\vspace{-0.4cm}
\caption{Accuracy and memory footprint of a 400-neurons U-SNN with different precision levels on the MNIST dataset \cite{Ref_Lecun_MNIST_IEEE98}. 
The W(\texttt{X})-N(\texttt{Y}) denotes \texttt{X} precision format for the weights and \texttt{Y} precision format for the neuron parameters (i.e., neurons' membrane and threshold potentials).} 
% These results show that different combinations of precision levels may achieve comparable accuracy at the training and the test phases.}
\label{Fig_ObserveHeteroQuant}
\vspace{-0.6cm}
\end{figure}

We observe that, apart from the weights, there are other SNN parameters that can be quantized to further reduce the memory footprint, 
% as they need to be stored in the on-chip memory \cite{Ref_Sen_ApproxSNN_DATE17}\cite{Ref_Roy_PEASE_ISLPED17}, 
e.g., neurons' membrane and threshold potentials (discussed further in Section~\ref{Sec_Background_SNNs}). 
To see the potential of such an idea, we study the impact of different \textit{precision levels} (bitwidth) for different SNN parameters on the accuracy through experiments using PyTorch-based simulation on GPGPU, i.e., Nvidia RTX 2080 Ti (the detailed experimental setup is explained in Section~\ref{Sec_Eval}). 
Fig.~\ref{Fig_ObserveHeteroQuant} shows the experimental results, from which we make the following key observations.
  \begin{itemize}[leftmargin=*]
    \item Different parameters may have different integer bitwidth, as they have different range of values. 
    \item Different combinations of precision levels (bitwidth) may achieve comparable accuracy, but occupy different memory footprint. For instance, W(\texttt{Q1.16})-N(\texttt{FP32}), W(\texttt{FP32})-N(\texttt{Q11.16}), and W(\texttt{Q1.16})-N(\texttt{Q11.16}) obtain about 84\% accuracy, while consuming ca. 1.2MB, 0.68MB, 1.19MB, and 0.67MB respectively; see label-\circledB{1} in Fig.~\ref{Fig_ObserveHeteroQuant}. 
    \item Less memory footprint requires less number of memory accesses, and thereby less access energy. 
    This potentially improves the energy-efficiency of SNN processing, as the memory accesses dominate the energy of SNN processing (i.e., 50\%-75\% of total system energy) \cite{Ref_Krithivasan_SpikeBundle_ISLPED19}.   
 \end{itemize}
Although the quantization effectively reduces the memory footprint, it leads to accuracy degradation if the quantization process is not performed carefully. 
Furthermore, finding the appropriate quantization levels for different SNN parameters is challenging, as the number of potential combinations of precision levels is large.
Therefore, the key challenge is \textit{how to effectively perform quantization and exploit the trade-off between memory and accuracy, so that the memory footprint is reduced and the targeted accuracy are met.}

%%%%%%%%%%%%%%%%%
%%%%%%%%%%%%%%%%%
\subsection{Our Novel Contributions}
\label{Sec_Novelty}
\vspace{-0.1cm}

\begin{figure}[hbtp]
\vspace{-0.2cm}
\centering
\includegraphics[width=\linewidth]{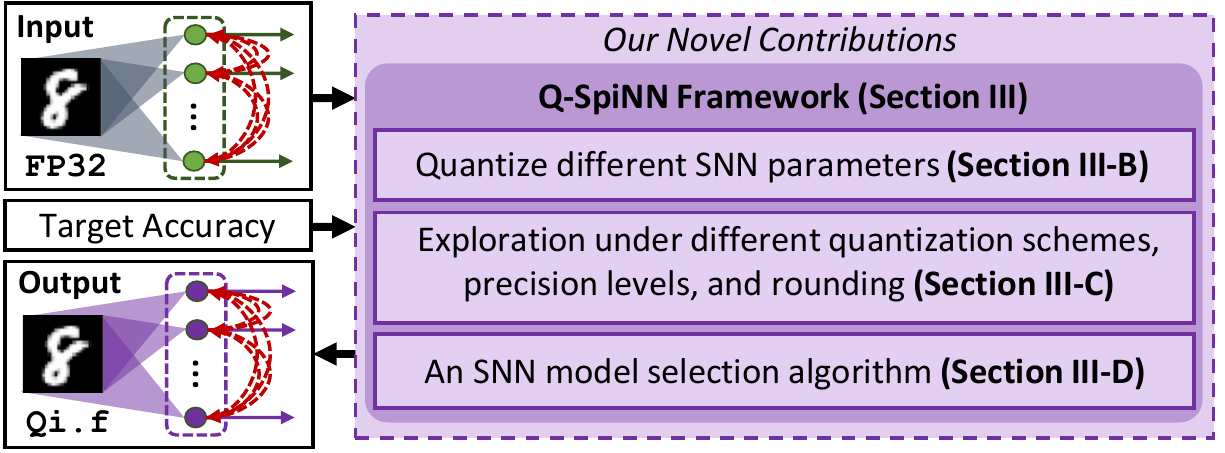}
\vspace{-0.6cm}
\caption{Overview of our novel contributions (shown in the purple boxes).}
\label{Fig_Novelty}
\vspace{-0.3cm}
\end{figure}

To address the above challenges, we propose \textbf{Q-SpiNN}, a novel \underline{Q}uantization framework for \underline{Spi}king \underline{N}eural \underline{N}etworks, through the following mechanisms (the overview is in Fig.~\ref{Fig_Novelty}).
\begin{itemize}[leftmargin=*]
    \item \textbf{Employ quantization for different SNN parameters} based on their significance to the accuracy, that are analyzed by observing the accuracy obtained under different precision.
    \item \textbf{Explore different combinations of quantization schemes, precision levels, and rounding schemes} to find the SNN models that meet the user-targeted accuracy, and refer them to as the solution candidates.  
    \item \textbf{Develop and employ an algorithm to select the SNN model from the given candidates}. 
    It quantifies the benefit of the memory-accuracy trade-off obtained by the candidates using the proposed reward function, and then selects the one with the highest benefit.
\end{itemize}

\textbf{Key Results:}
We evaluated the Q-SpiNN using PyTorch- based simulation on the GPGPU and the Embedded GPU.
The experimental results show that, for the unsupervised SNN, Q-SpiNN achieves 4x memory saving, while maintaining the accuracy within 1\% from the baseline on the MNIST. 
For the supervised one, it achieves 2x memory saving, with the accuracy within 2\% from the baseline on the DVS-Gesture. 

%%%%%%%%%%%%%%%%%
%%%%%%%%%%%%%%%%%
% \subsection{Paper Organization}
% \label{Sec_Intro_PaperOrg}

% Section~\ref{Sec_Background} presents the background of SNNs and Section~\ref{Sec_QSpiNN} discusses the Q-SpiNN framework. 
% We explain the evaluation methodology in Section~\ref{Sec_Eval}. 
% Then, we present the results and discussion in Section~\ref{Sec_Results}. 
% Section~\ref{Sec_Conclusion} concludes the paper.

%%%%%%%%%%%%%%%%%%%%%%%%%%%%%%%%%%%%%%%%%%%%%%%%%%%%%%%%%%%%%%%%%%%%%%%%%%%%%%%%%%%%%%%%%%%%%%%%%%%%%%%%%%%%%%%%%%%%%%%%%%%%%%%%%%%%%%%%%%%%%%%%%%%%%%%%%%%%%%%%%%%%%%%%
\section{Background and Related Work}
\label{Sec_Background}
\renewcommand{\headrulewidth}{0pt}
% \vspace{-0.1cm}

%%%%%%%%%%%%%%%%%
%%%%%%%%%%%%%%%%%
\subsection{Spiking Neural Networks (SNNs)}
\label{Sec_Background_SNNs}
% \vspace{-0.1cm}

An SNN model is composed of the network architecture, the neuron and synapse model, the spike coding, and the learning rule \cite{Ref_Mozafari_SpykeTorch_FNINS19}.
There are two major learning approaches that determine how the SNN models are designed and trained, i.e., the unsupervised learning and the supervised learning.
\textit{In this paper, we evaluate our Q-SpiNN framework for both learning approaches to show its generality for different SNN designs.} 
For the unsupervised SNN, we consider the single-layer network that employs the spike-timing-dependent plasticity (STDP) learning \cite{Ref_Putra_FSpiNN_TCAD20}.
For the supervised one, we consider the multi-layer network that employs the deep continuous local learning (DECOLLE) \cite{Ref_Kaiser_DECOLLE_FNINS20}.
We select them since they show the state-of-the-art accuracy with relatively low memory and compute costs, compared to other designs with same approach.

\smallskip
\textbf{A Single-Layer SNN with Unsupervised Local Learning (U-SNN):} 
This network consists of a single fully-connected (FC) layer. 
Each input pixel is converted into the rate-coded spikes which are transferred to all neurons. 
% and the connecting synapses learn the input features based on the learning rule.
Each neuron generates spikes that inhibit other neurons, thereby enabling competition among neurons, as shown in Fig.~\ref{Fig_ObserveQuant}(a). 
Here, the pair-wise weight-dependent STDP learning rule is used, as it defines the maximum allowed weights, which is suitable for fixed-point format (see Eq.~\ref{Eq_PairWeightSTDP}). 
\begin{equation}
\vspace{-0.3cm}
\small
\begin{split}
\Delta w = 
\begin{cases}
-\eta_{pre} x_{post} w^\mu & \text{on} \; \text{presynaptic spike}\\
\eta_{post} x_{pre} (w_{m}-w)^\mu & \text{on} \; \text{postsynaptic spike}
\end{cases}
\label{Eq_PairWeightSTDP}
\end{split}
% \vspace{0.3cm}
\end{equation}
\vspace{0.1cm}
\\
$\Delta w$ denotes the weight update, $\eta_{pre}$ and $\eta_{post}$ denote the learning rate for pre- and post-synaptic spike, while $x_{pre}$ and $x_{post}$ denote the pre- and post-synaptic traces, respectively.
$w_m$ denotes the maximum allowed weight, $w$ denotes the current weight, and $\mu$ denotes the weight dependence factor.

Here, the conductance-based Leaky Integrate-and-Fire (LIF) neuron model is used, since it has low complexity \cite{Ref_Izhikevich_CompareModels_TNN04}. 
Its membrane potential ($V$) increases each time a presynaptic spike comes, otherwise it decreases. % exponentially. 
If the $V$ reaches the threshold potential ($V_{th}$), a spike is emitted, then it goes to the reset potential ($V_{reset}$).
To prevent a neuron from dominating the spike firing, the $V_{th}$ is defined as $V_{th}+\theta$ with $\theta$ refers to as the adaptation potential, which adds to $V_{th}$ each time the neuron fires a spike. 
% and otherwise $V_{th}$ decays with a $\theta_{decay}$ rate. 
% \footnote{For the detailed discussion on the mechanism of this architecture, we refer to the paper \cite{Ref_Putra_FSpiNN_TCAD20}.}.
% This behavior is known as \textit{homeostasis} which prevents a neuron from dominating the spike firing.
The synapse is modeled by a conductance, which increases by weight ($w$) when a presynaptic spike comes, and otherwise it decreases. 
\textit{Note, we quantize these weights and neuron parameters to get memory saving.}

\begin{figure}[hbtp]
% \vspace{-0.3cm}
\centering
\includegraphics[width=0.93\linewidth]{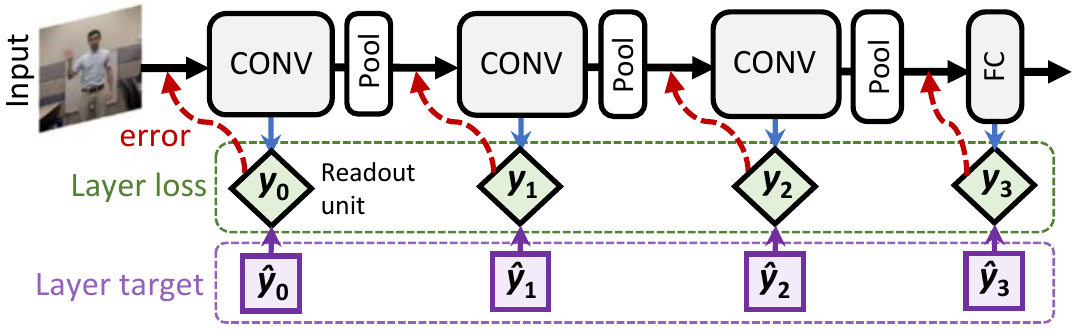}
\vspace{-0.4cm}
\caption{The multi-layer network with supervised local learning, which is adapted from the studies in \cite{Ref_Kaiser_DECOLLE_FNINS20}.}
\label{Fig_DecolleNetwork}
\vspace{-0.6cm}
\end{figure}

\smallskip
\textbf{A Multi-Layer SNN with Supervised Local Learning (S-SNN):}
This network consists of three convolutional (CONV) layers and one FC layer, as shown in Fig.~\ref{Fig_DecolleNetwork}. 
Each layer is trained using the supervised deep continuous local learning (DECOLLE) \cite{Ref_Kaiser_DECOLLE_FNINS20}, whose idea is to use a surrogate gradient for minimizing the local (layer-wise) loss function, so that the readout unit can produce the targeted output ($\hat{y}$). 
The difference between the readout output ($y$) and the target ($\hat{y}$) denotes the error that is used to train the weights (red dashed-line). 
In this manner, the loss function minimization can be performed directly in the spiking environment.

The dynamics of each layer are based on the current-based LIF neuron model, and expressed as Eq.~\ref{Eq_NeuronSynapse}. 
$V_i^l[n]$ denotes the membrane potential of neuron-$i$ in layer-$l$ at timestep-$n$, while $w_{ij}$ denotes the weight between the pre-synaptic neuron-$j$ and the post-synaptic neuron-$i$.
A spike $S_i^l[n]$ is emitted at timestep-$n$ if $V_i^l[n]$ reaches the threshold ($V_{th}$) through the $\Theta$ function, where $\Theta(x) = 1$ if $x \geq 0$, and otherwise 0. 
$P$ and $Q$ denote the traces of the membrane and the current-based synapse respectively, while $R$ denotes the refractory state and $\rho$ is the inhibition weight.
$\alpha = exp(-\frac{\Delta t}{\tau_{mem}})$, $\beta = exp(-\frac{\Delta t}{\tau_{syn}})$, and $\gamma = exp(-\frac{\Delta t}{\tau_{ref}})$ denote the decay of the $V$, $Q$, and $R$, respectively. 
For the detailed discussion on the DECOLLE, we refer to the original paper \cite{Ref_Kaiser_DECOLLE_FNINS20}. 
\textit{We quantize these weights and neuron parameters to get memory saving.}

\vspace{-0.3cm}
\begin{equation}
\vspace{-0.4cm}
\small
\begin{split}
V_i^l[n] & = \sum\limits_j w_{ij}^l P_j^l[n] - \rho R_i^l[n] \\ 
S_i^l[n] & = \Theta (V_i^l[n] - V_{th}) \\
P_j^l[n+1] & = \alpha P_j^l[n] + Q_j^l[n] \\
Q_j^l[n+1] & = \beta Q_j^l[n] + S_j^{l-1}[n] \\
R_i^l[n+1] & = \gamma R_i^l[n] + S_i^l[n] \\
\label{Eq_NeuronSynapse}
\end{split}
\vspace{-0.6cm}
\end{equation}
\vspace{-0.3cm}

%%%%%%%%%%%%%%%%%
%%%%%%%%%%%%%%%%%
% \vspace{-0.1cm}
\subsection{Fixed-Point Representation and Rounding Schemes}
\label{Sec_Background_FixedPoint}
\vspace{-0.1cm}

The fixed-point format is represented as \texttt{Qi.f}, that consists of 1 sign bit, \texttt{i} integer bits, and \texttt{f} fractional bits, and follows the 2's complement format. 
Given the fixed-point \texttt{Qi.f}, the range of representable values is $[-2^{\texttt{i}}, 2^{\texttt{i}}-2^{\texttt{-f}}]$ and the precision is $\epsilon = 2^{\texttt{-f}}$. 
%
%%%%%%%
In the quantization process, a \textit{rounding scheme} is required, and we consider the widely used ones, i.e., truncation, rounding-to-the-nearest, and stochastic \cite{Ref_Hopkins_Rounding_RSTA20} \cite{Ref_Gupta_DLPrecision_ICML15}.

\textbf{Truncation (TR)} 
keeps the \texttt{f} bits and discards the other bits from the fractional part. 
Hence, the output fixed-point for the given real number $x$ and configuration \texttt{Qi.f}, is defined as $TR(x,\texttt{Qi.f}) = \lfloor x \rfloor$.

\textbf{Rounding-to-the-Nearest (RN)}
rounds the value, that is half-way between two representable values ($\lfloor x \rfloor + \frac{\epsilon}{2}$), by rounding it up. 
Hence, the output fixed-point for the given real number $x$ and configuration \texttt{Qi.f}, is defined as 
\begin{equation}
% \vspace{-0.2cm}
\small
\begin{split}
RN(x,\texttt{Qi.f}) = 
\begin{cases}
\lfloor x \rfloor & \text{if} \; \lfloor x \rfloor \leq x < \lfloor x \rfloor + \displaystyle \frac{\epsilon}{2}\\
\lfloor x \rfloor + \epsilon & \text{if} \; \lfloor x \rfloor + \displaystyle \frac{\epsilon}{2} \leq x < \lfloor x \rfloor + \epsilon
\end{cases}
\label{Eq_RN}
\end{split}
\vspace{-0.2cm}
\end{equation}

\textbf{Stochastic Rounding (SR)}
rounds the value using a non- deterministic approach.  
Given a random value $P \in [0,1)$ that is drawn from a uniform random number generator, the output fixed-point for the real number $x$ and configuration \texttt{Qi.f} is % defined as
\begin{equation}
% \vspace{-0.2cm}
\small
\begin{split}
SR(x,\texttt{Qi.f}) = 
\begin{cases}
\lfloor x \rfloor & \text{if} \; P \geq \displaystyle \frac{x-\lfloor x \rfloor}{\epsilon}  \\
\lfloor x \rfloor + \epsilon & \text{if} \; P < \displaystyle \frac{x-\lfloor x \rfloor}{\epsilon}
\end{cases}
\label{Eq_SR}
\end{split}
\vspace{-0.1cm}
\end{equation}

%%%%%%%%%%%%%%%%%
%%%%%%%%%%%%%%%%%
\vspace{-0.1cm}
\subsection{Quantization Schemes}
\label{Sec_Background_QuantScheme}
\vspace{-0.1cm}

\begin{figure}[hbtp]
\vspace{-0.3cm}
\centering
\includegraphics[width=\linewidth]{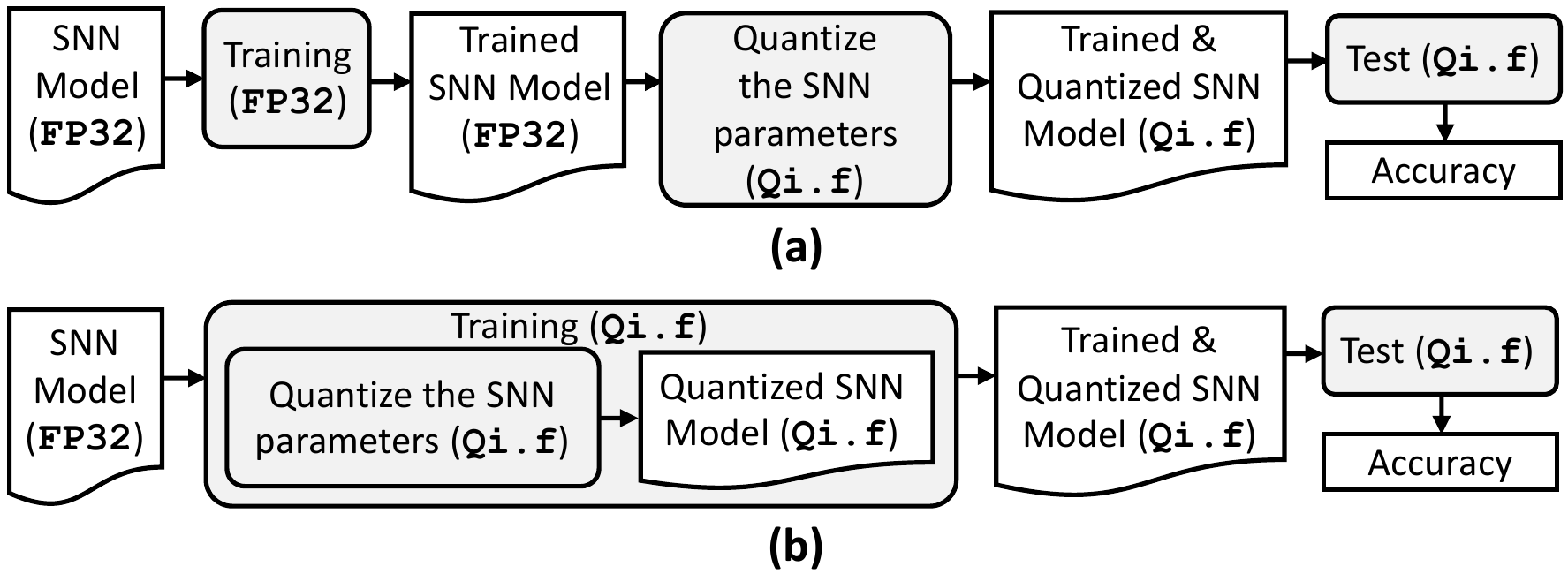}
\vspace{-0.7cm}
\caption{Overview of (a) the post-training quantization, and (b) the in-training quantization or quantization-aware training.}
\label{Fig_QuantSchemes}
% \vspace{-0.4cm}
\end{figure}

There are two widely used quantization schemes in the neural network models, i.e., the \textit{Post-Training Quantization}, and the \textit{In-Training Quantization} (or the \textit{Quantization-aware Training}) \cite{Ref_Krishnamoorthi_Whitepaper_arXiv18}, whose key mechanisms are shown in Fig.~\ref{Fig_QuantSchemes}.
% Overview of their mechanisms are shown in Fig.~\ref{Fig_QuantSchemes}.

\textbf{Post-Training Quantization (PTQ)} 
trains an SNN model with a floating-point precision (e.g., \texttt{FP32}) and results in a trained model. 
Afterwards, the quantization is performed on the trained model with the given \texttt{Qi.f} precision, resulting in a quantized model for the inference phase. 

\textbf{In-Training Quantization (ITQ)} 
quantizes an SNN model with the given \texttt{Qi.f} precision during the training phase. 
Therefore, the trained model is already in a quantized form and can be used for the inference phase. 
The quantization is typically performed using the simulated quantization \cite{Ref_Krishnamoorthi_Whitepaper_arXiv18}.

\vspace{0.4cm}

%%%%%%%%%%%%%%%%%%%%%%%%%%%%%%%%%%%%%%%%%%%%%%%%%%%%%%%%%%%%%%%%%%%%%%%%%%%%%%%%%%%%%%%%%%%%%%%%%%%%%%%%%%%%%%%%%%%%%%%%%%%%%%%%%%%%%%%%%%%%%%%%%%%%%%%%%%%%%%%%%%%%%%%%
\section{Our Q-SpiNN Framework}
\label{Sec_QSpiNN}

%%%%%%%%%%%%%%%%%
%%%%%%%%%%%%%%%%%
\vspace{-0.1cm}
\subsection{Overview}
\label{Sec_QSpiNN_Overview}
\vspace{-0.1cm}

\begin{figure}[hbtp]
\vspace{-0.4cm}
\centering
\includegraphics[width=0.9\linewidth]{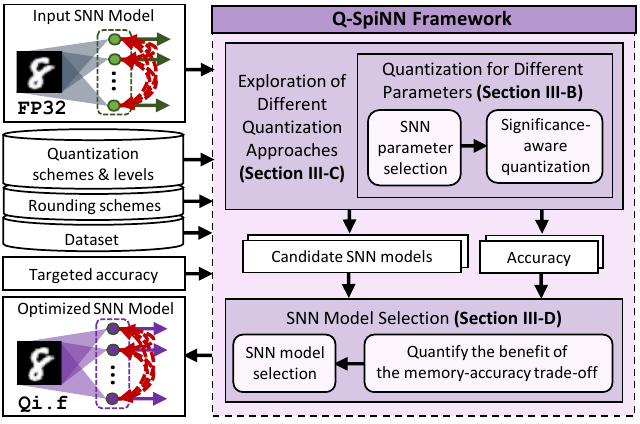}
\vspace{-0.3cm}
\caption{The detailed mechanisms of our Q-SpiNN framework. The novel steps are shown in purple boxes.}
\label{Fig_QSpiNN}
\vspace{-0.4cm}
\end{figure}

The Q-SpiNN framework employs the following key mechanisms for obtaining memory-efficient SNNs, while maintaining the accuracy (the overview is shown in Fig.~\ref{Fig_QSpiNN}). 

\begin{itemize}[leftmargin=*]
    \item \textbf{Quantization of different parameters (\textbf{Section~\ref{Sec_QSpiNN_Quant}}): } 
    It is performed through the following means. 
      \begin{itemize}
          \item Maximizing the quantization for each SNN parameter. 
          \item Defining the precision level (bitwidth) for each parameter based on its significance, that is obtained by analyzing the accuracy under different precision levels. 
      \end{itemize}
    \item \textbf{Design exploration of different quantization approaches (\textbf{Section~\ref{Sec_QSpiNN_DSE}}):} 
    It is done through the following means. 
      \begin{itemize}
          \item Observing the accuracy obtained by different quantization schemes (i.e., PTQ and ITQ), different precision levels, and different rounding schemes (i.e., TR, RN, and SR), 
          \item Selecting the SNN models that meet the targeted accuracy as the solution candidates.
      \end{itemize}
    \item \textbf{An SNN model selection  (\textbf{Section~\ref{Sec_QSpiNN_Select}}):} 
    It searches for an appropriate SNN model from the given candidates through the following means.
      \begin{itemize}
          \item Quantifying the benefit of the memory-accuracy trade-off obtained by the SNN model candidates using \textit{our proposed multi-objective reward function},
          \item Selecting the SNN model with the highest benefit.
      \end{itemize}
\end{itemize}

%%%%%%%%%%%%%%%%%
%%%%%%%%%%%%%%%%%
\vspace{-0.2cm}
\subsection{Quantization of Different SNN Parameters}
\label{Sec_QSpiNN_Quant}
\vspace{-0.1cm}

\begin{figure}[hbtp]
\vspace{-0.3cm}
\centering
\includegraphics[width=\linewidth]{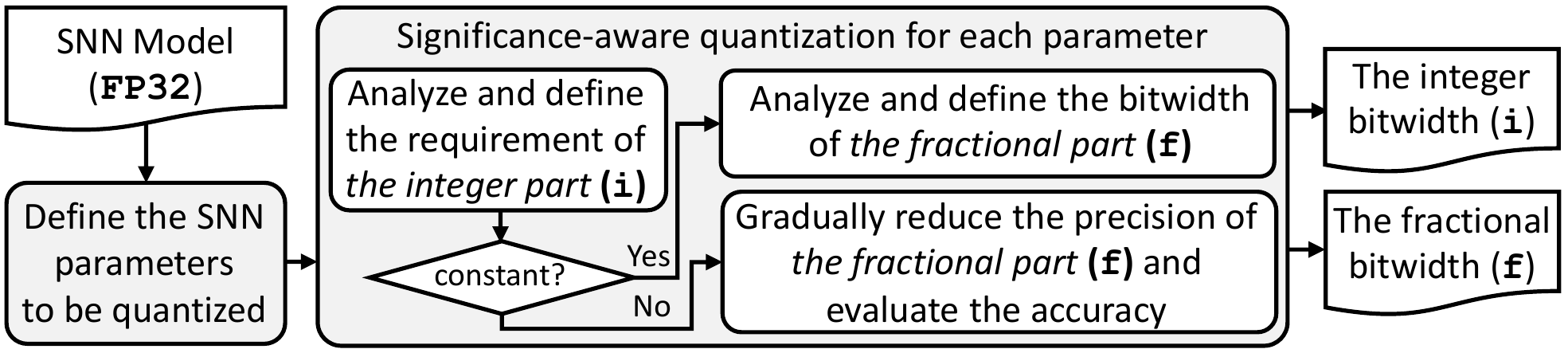}
\vspace{-0.7cm}
\caption{Overview of the proposed quantization steps for the given SNN model.}
\label{Fig_Quantize}
\vspace{-0.3cm}
\end{figure}

Different SNN designs may have different SNN parameters that can be quantized, as discussed in Section~\ref{Sec_Background_SNNs}. 
Therefore, \textit{to provide a generic solution for any SNN designs, we propose a significance-aware quantization steps} (overview is in Fig.~\ref{Fig_Quantize}).  
The idea is to maximize the quantization for each SNN parameter, and define the precision level for each parameter based on its significance to the accuracy. 
For the given SNN model (in \texttt{FP32}), we first determine the parameters to be quantized by manually selecting them.
Afterward, we analyze the significance of each parameter to determine the integer and fractional bitwidth. 
For the integer part, the bitwidth requirement is analyzed by observing the range of parameter values when running the given workload. 
For the fractional part, there are two cases. 
If the parameter is a constant, then the bitwidth depends on the parameter value; and otherwise (if parameter is a variable), the bitwidth requirement is analyzed by gradually reducing its precision and observing the output accuracy. 
In this manner, the impact of the parameters' bitwidth on accuracy is systematically explored. 

\smallskip
\textbf{Case Study:} 
We provide a case study to show how the proposed quantization steps are done for the unsupervised SNN (U-SNN) with MNIST. 
First, we select the $w$, $V$, $V_{th}$, $V_{reset}$, and $\theta$ as the parameters to be quantized (see SNN fundamentals in Section~\ref{Sec_Background_SNNs}). 
$V_{reset}$ and $\theta$ are \textit{constants}, while others (i.e., $w$, $V$, and $V_{th}$) are \textit{variables} in the training. 
\begin{itemize}[leftmargin=*]
    \item \textit{For constant parameters:} 
    We quantize the constants based on their values (see Table~\ref{Table_USNN_Quant}). 
    For instance, $V_{reset}$ = -65mV and it is represented using 8 bits with \texttt{Q7.0} format, 
    while $\theta$ = 0.05mV and it is represented using 8 bits with \texttt{Q1.6} format. 
    In this manner, 24 bits are saved for $V_{reset}$ and $\theta$, compared to the \texttt{FP32}.
    \item \textit{For variable parameters:}
    We perform experiments to obtain the ranges of parameter values (see the values in Table~\ref{Table_USNN_Quant}).
    \begin{itemize}
        \item For the integer part, we define the integer bitwidth based on the observed ranges, i.e., $V_{th}$, $V$, and $w$ need 11 bits, 11 bits, and 1 bit of integer, respectively.
        \item For the fractional part, we gradually reduce the precision and observing the output accuracy to study the impact of different precision levels. 
        Therefore, \textit{we perform a design exploration}, which is discussed further in Section~\ref{Sec_QSpiNN_DSE}.
    \end{itemize}
\end{itemize}

%%%%%
\vspace{-0.3cm}
\begin{table}[hbtp]
\vspace{-0.2cm}
\caption{Observed U-SNNs' Parameter Values for MNIST Workload.}
\label{Table_USNN_Quant}
\vspace{-0.2cm}
\centering
\scriptsize
\begin{tabular}{|c|c|c|}
\hline 
\multirow{1}{*}{\textbf{Parameters}} & \multirow{1}{*}{\textbf{Value}} & \multirow{1}{*}{\textbf{Description}} \\ 
\hline
\hline
$V_{reset}$ & -65mV & shown by label-\circled{1} in Fig.~\ref{Fig_USNN_QuantVar}(a) \\
\hline
$\theta$ & 0.05mV & shown by label-\circled{2} in Fig.~\ref{Fig_USNN_QuantVar}(a) \\
\hline
$V_{th}$ & -52mV -- 1271.88mV & shown by label-\circled{3} in Fig.~\ref{Fig_USNN_QuantVar}(a) \\
\hline
$V$ & -887.29mV -- 1250.18mV & shown by label-\circled{4} in Fig.~\ref{Fig_USNN_QuantVar}(a) \\
\hline
$w$ & 0 -- 0.7 & shown by label-\circled{5} in Fig.~\ref{Fig_USNN_QuantVar}(b) \\
\hline
\end{tabular}
\vspace{-0.3cm}
\end{table}
%%%%%%

\begin{figure}[hbtp]
\vspace{-0.4cm}
\centering
\includegraphics[width=0.95\linewidth]{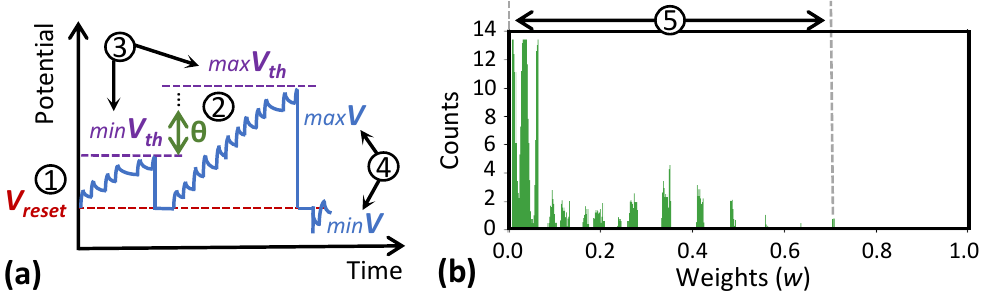}
\vspace{-0.4cm}
\caption{(a) Overview how the neuron parameters are involved in the neuron dynamics. (a) The weight distribution of the trained U-SNN model with 400 excitatory neurons.}
\label{Fig_USNN_QuantVar}
\vspace{-0.4cm}
\end{figure}

%%%%%%%%%%%%%%%%%
%%%%%%%%%%%%%%%%%
\vspace{-0.1cm}
\subsection{Exploration of Different Quantization Approaches}
\label{Sec_QSpiNN_DSE}
\vspace{-0.1cm}

To find the effective configuration of quantization for the given SNN, a design exploration of different quantization approaches is required.
Therefore, \textit{we comprehensively study the impact of different quantization schemes (i.e., PTQ and ITQ), different precision levels, and different rounding schemes (i.e., TR, RN, and SR), and select the ones that meet the user- targeted accuracy}. 
Towards this, we devise a search algorithm across a range of selected parameters to systematically perform the exploration, whose steps are the following (pseudo-code is in Alg.~\ref{Alg_Search}, considering an example for the U-SNN case). 

\begin{itemize}[leftmargin=*]
    \item We train the given model with a floating-point precision (Alg.~\ref{Alg_Search}: line 3), and the test accuracy of the trained model is considered as the baseline accuracy (Alg.~\ref{Alg_Search}: line 4). 
    Then, we perform the PTQ and the ITQ schemes, subsequently.
    \item For the PTQ, the quantization is performed on the trained model, then the accuracy is evaluated (Alg.~\ref{Alg_Search}: lines 12-16).
    Meanwhile for the ITQ, we quantize the given SNN model during the training phase. 
    Therefore, the trained model is already in a quantized form, and can be used for the accuracy evaluation (Alg.~\ref{Alg_Search}: lines 18-23). 
    % The quantization is performed using the simulated quantization operations \cite{Ref_Krishnamoorthi_Whitepaper_arXiv18}. 
    %
    \item For both schemes, we reduce the precision of each parameter using a nested for-loop (Alg.~\ref{Alg_Search}: lines 7-9), and in each step, we explore the use of different rounding schemes (Alg.~\ref{Alg_Search}: line 10).
    The depth of the loop depends on the parameters (e.g., we consider $w$, $V$, and $V_{th}$ for U-SNN case). 
    If the accuracy is within the target, then the model is selected as a solution candidate. Otherwise, the currently investigated precision and the lower precision (if any) for the corresponding parameter, are not considered in next exploration steps (Alg.~\ref{Alg_Search}: lines 24-33). 
    % (Alg.~\ref{Alg_Search}: lines 25-36). 
    \textit{Therefore, the design space is reduced and the exploration is performed efficiently.}
\end{itemize}
This exploration populates the SNN models that meet the target accuracy (i.e., the solution candidates). 
% and we refer them to as the solution candidates. 
To select the appropriate model out of them, we propose a model selection algorithm, which is discussed in the Section~\ref{Sec_QSpiNN_Select}.

\begin{algorithm}[t]
% \footnotesize
\scriptsize
\caption{Pseudo-code of the proposed exploration}
\label{Alg_Search}
\begin{algorithmic}[1]
\renewcommand{\algorithmicrequire}{\textbf{INPUT:}}
\renewcommand{\algorithmicensure}{\textbf{OUTPUT:}}
\REQUIRE \textbf{(1)} SNN: floating-point model  ($m_{fp}$), accuracy ($acc_{fp}$), parameters ($m_{fp}.w, m_{fp}.V$, and $m_{fp}.V_{th}$); // for the U-SNN case study; 
% variable parameters $w, V$, and $V_{th}$ are considered; 
\textbf{(2)} Maximum allowed accuracy degradation ($acc_{target}$);  
\textbf{(3)} Quantization schemes ($QS = [PTQ, ITQ]$); 
\textbf{(4)} Precision levels ($QL$); // $QL$ is the user-defined  fractional bitwidth sorted in a descending order, e.g., $QL = [16, 14, ..., 0]$; 
\textbf{(5)} Rounding schemes ($RS = [TR, RN, SR]$); \\
\ENSURE SNN model candidates ($C$); \\ 
\vspace{0.1cm}
\renewcommand{\algorithmicrequire}{\textbf{BEGIN}}
\renewcommand{\algorithmicensure}{\textbf{END}}
\REQUIRE \hspace{0.1cm} \\   
\textbf{Initialization}: \\
  \STATE $C = []$; \\
  \STATE $c = 1$; \\
\textbf{Process}: \\
  \STATE $\overline{m}_{fp} \leftarrow train (m_{fp})$; \\
  \STATE $acc_{fp} = test (\overline{m}_{fp})$; \\
  \FOR{($qs = 1$ to $len(QS)$)}
    \STATE $Nw = len(QL)$; $Nv = len(QL)$; $Nt = len(QL)$; \\
    % \STATE $Nv = len(QL)$; \\
    % \STATE $Nt = len(QL)$; \\
  \FOR{($iw = 1$ to $Nw$)}
    \FOR{($iv = 1$ to $Nv$)}
      \FOR{($it = 1$ to $Nt$)}
        \FOR{($rs = 1$ to $len(RS)$)}
          \IF{($QS[qs] == PTQ$)}
            \STATE $w_q = quantize(\overline{m}_{fp}.w,QL[iw],RS[rs])$; 
            \STATE $V_q = quantize(\overline{m}_{fp}.V,QL[iv],RS[rs])$;
            \STATE $V_{thq} = quantize(\overline{m}_{fp}.V_{th},QL[it],RS[rs])$;
            \STATE $\overline{m}_q \leftarrow substitute(\overline{m}_{fp}, (w_q, V_q, V_{thq}))$;\\
            \STATE $acc_{q} = test (\overline{m}_{q})$; \\
          \ELSE 
            \STATE $w_q = quantize(m_{fp}.w,QL[iw],RS[rs])$; 
            \STATE $V_q = quantize(m_{fp}.V,QL[iv],RS[rs])$;
            \STATE $V_{thq} = quantize(m_{fp}.V_{th},QL[it],RS[rs])$;
            \STATE $m_q \leftarrow substitute(m_{fp}, (w_q, V_q, V_{thq}))$;\\
            \STATE $\overline{m}_q \leftarrow train (m_q)$; \\
            \STATE $acc_{q} = test (\overline{m}_{q})$; \\
          \ENDIF  
            \IF{$(acc_q \geq (acc_{fp}-acc_{target}))$}
              \STATE $C[c] = \overline{m}_q$; \\
              \STATE $c \mathrel{+}= 1$; \\
            \ELSE
              \IF{$(iw \geq 1) \& (iv == 1) \& (it == 1)$}
                \STATE $Nw = iw-1$;
              \ELSIF{$(iw == 1) \& (iv \geq 1) \& (it == 1)$}
                \STATE $Nv = iv-1$;
              \ELSIF{$(iw == 1) \& (iv == 1) \& (it \geq 1)$}
                \STATE $Nt = it-1$;
              \ENDIF
            \ENDIF
        \ENDFOR
      \ENDFOR
    \ENDFOR
  \ENDFOR 
  \ENDFOR 
\RETURN $C$;
\ENSURE
\end{algorithmic} 
\end{algorithm}
\setlength{\textfloatsep}{1pt}

%%%%%%%%%%%%%%%%%
%%%%%%%%%%%%%%%%%
\vspace{-0.1cm}
\subsection{SNN Model Selection Algorithm}
\label{Sec_QSpiNN_Select}
\vspace{-0.1cm}

We obtain a set of model candidates 
% with their corresponding accuracy and quantization 
from the exploration in Section~\ref{Sec_QSpiNN_DSE}. 
Afterwards, we need to select the Pareto-optimal model out of the candidates, while considering the accuracy and the memory footprint.
Towards this, \textit{we propose an SNN model selection algorithm that quantifies the benefit of the memory-accuracy trade-off obtained by the candidates using the proposed multi-objective reward function}. 
The idea of our reward function is to prioritize the model with higher accuracy and smaller memory footprint, which is expressed as Eq.~\ref{Eq_Reward}. 
% Therefore, applying this function to all candidates will identify the suitable SNN model for the given design requirements.
%
\vspace{-0.1cm}
\begin{equation}
\small
R(acc_q,m_{norm}) = acc_q - \mu \; m_{norm}
\label{Eq_Reward}
\end{equation}
\begin{equation}
\small
m_{norm} = \frac{mem_q}{mem_{fp}}
\label{Eq_NormMemory}
\end{equation}
\begin{equation}
\small
\begin{split}
mem &= mem\_w + mem\_n = Nw \; Bw + \sum_k Nn^k \; Bn^k 
\end{split}
\label{Eq_Memory}
\vspace{-0.1cm}
\end{equation}
$acc_q$ denotes the test accuracy of the quantized SNN model, $m_{norm}$ denotes the normalized memory footprint, and the coefficient $\mu$ is the weight to trade-off between memory and accuracy.
Note $acc$, $m_{norm}$, and $\mu$ have a value range of [0,1].
$m_{norm}$ is obtained from the ratio between the memory of quantized model ($mem_q$) and floating-point model ($mem_{fp}$), as stated in Eq.~\ref{Eq_NormMemory}.
The memory footprint ($mem$) is estimated by the total memory required by the weights ($mem\_w$) and neuron parameters ($mem\_n$), as shown in Eq.~\ref{Eq_Memory}.  
$mem\_w$ is obtained by multiplying the number of weights ($Nw$) and the respective bitwidth ($Bw$).
Similar approach is used for neuron parameters, i.e., multiplying the number of parameter ($Nn$) and the bitwidth ($Bn$). 
Since the neuron has several parameters ($k$) which may have different precision, $mem\_n$ is defined as the total bits from all neuron parameters.

%%%%%%%%%%%%%%%%%%%%%%%%%%%%%%%%%%%%%%%%%%%%%%%%%%%%%%%%%%%%%%%%%%%%%%%%%%%%%%%%%%%%%%%%%%%%%%%%%%%%%%%%%%%%%%%%%%%%%%%%%%%%%%%%%%%%%%%%%%%%%%%%%%%%%%%%%%%%%%%%%%%%%%%%
\vspace{-0.2cm}
\section{Evaluation Methodology}
\label{Sec_Eval}
\vspace{-0.2cm}

\begin{figure}[t]
% \vspace{-0.3cm}
\centering
\includegraphics[width=0.88\linewidth]{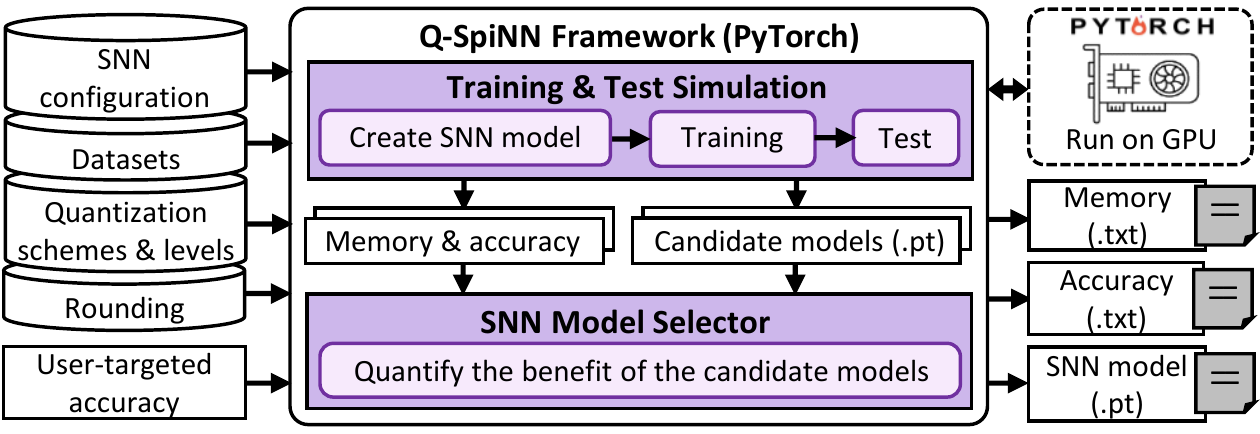}
\vspace{-0.4cm}
\caption{Experimental setup for evaluating our Q-SpiNN framework.}
\label{Fig_ExpSetup}
% \vspace{-0.1cm}
\end{figure}

Fig.~\ref{Fig_ExpSetup} shows the experimental setup for evaluating the Q-SpiNN framework. 
We use the PyTorch-based simulation to evaluate the accuracy of the unsupervised SNN \cite{Ref_Hazan_BindsNET_FNINF18} and the supervised SNN \cite{Ref_Kaiser_DECOLLE_FNINS20}, estimate the memory, and select the SNN model. 
We run the simulations on GPGPU (i.e., Nvidia GeForce RTX 2080 Ti) and Embedded GPU (i.e., Nvidia Jetson Nano) to show the applicability of the Q-SpiNN framework on different hardwares with different compute and memory capabilities. 

\textbf{Networks:} 
We use networks with different architectures, number of layers, and learning rules to show the generality of our Q-SpiNN. 
For the unsupervised SNN, we consider a single-layer FC network with the STDP, as shown in Fig.~\ref{Fig_ObserveQuant}(a), while for the supervised SNN, we consider a multi-layer CONV network with the DECOLLE, as shown in Fig.~\ref{Fig_DecolleNetwork}.   

\textbf{Datasets:} 
We use the MNIST dataset \cite{Ref_Lecun_MNIST_IEEE98} for the U-SNN, and the DVS-Gesture dataset \cite{Ref_Amir_DvsGesture_CVPR17} for the S-SNN. 
In the MNIST, there are 60,000 images for the training and 10,000 images for the test, each having a dimension of 28×28 pixels.
Meanwhile, the DVS-Gesture, which is obtained using a Dynamic Vision Sensor (DVS), consists of 1,342 instances of a set of 11 hand and arm gestures.
They are collected from 29 subjects under 3 lighting conditions. 
Gestures from 23 subjects are used as the training set, and the remaining 6 subjects are used as test set.
Each gesture consists of the stream of events and lasts for 6s. % and has unique features within the stream of events.  
The event streams were downsized from 128 × 128 to 32 × 32 
% (by summing the events from 4 neighboring pixels as a common stream) 
and binned in frames of 1ms \cite{Ref_Kaiser_DECOLLE_FNINS20}. 
% as the timestep for the GPU-based simulation \cite{Ref_Kaiser_DECOLLE_FNINS20}. 
%
% The DvsGesture dataset comprises 1,342 instances of a set of 11 hand and arm gestures (Fig. 7), grouped in 122 trials collected from 29 subjects under 3 different lighting conditions. During each trial one subject stood against a stationary background and performed all 11 gestures sequentially under the same lighting condition. The gestures include hand waving (both arms), large straight arm rotations (both arms, clockwise and counterclockwise), forearm rolling (forward and backward), air guitar, air drums, and an “Other” gesture invented by the subject. The 3 lighting conditions are combinations of natural light, fluorescent light, and LED light, which were selected to control the effect of shadows and fluorescent light flicker on the DVS128. Each gesture lasts about 6 seconds. To evaluate classifier performance, 23 subjects are designated as the training set, and the remaining 6 subjects are reserved for out-of-sample validation.

\textbf{Comparisons:} 
We use networks with different precision levels as the comparison partners, for both the unsupervised (i.e., U-SNN) and supervised (i.e., S-SNN) cases. 
For the U-SNN, we consider a network with 400 excitatory neurons with 1 training epoch (i.e., using the STDP during forward propagation). 
For the S-SNN, we train the network using the DECOLLE with 200 epochs. 
For both cases, the baseline refers to the network with \texttt{FP32} precision for all parameters. 

\textbf{Quantization Format:}
We use the W(\texttt{X})-N(\texttt{Y}) format to represent a model with \texttt{X} precision for the weights and \texttt{Y} precision for the neuron parameters (see Section~\ref{Sec_QSpiNN_Quant}). 
For conciseness, we simply use W(\texttt{Qi.f}) to represent a model with W(\texttt{Qi.f})-N(\texttt{FP32}) precision, and N(\texttt{Qi.f}) to represent a model with W(\texttt{FP32})-N(\texttt{Qi.f}) precision. 
Furthermore, since there are several neuron parameters involved in the quantization process, their integer part is simply written as \texttt{i}, e.g., N(\texttt{Qi.8}) means that each neuron parameter employs integer bitwidth based on its value range and 8-bit fraction.

% \begin{table}[t]
% \vspace{-0.2cm}
% \caption{GPU Specifications.}
% \label{Table_GPUspecs}
% \vspace{-0.2cm}
% \centering
% % \small
% \footnotesize
% \begin{tabular}{|c|c|c|}
% \hline 
% \textbf{Category} & \textbf{Jetson Nano} & \textbf{RTX 2080 Ti} \\
% \hline
% \hline
% Architecture & Maxwell & Turing \\
% \hline
% CUDA cores & 128 & 4352 \\
% \hline
% Memory & 4GB LPDDR4 & 11GB GDDR6 \\
% \hline
% Interface width & 64-bit & 352-bit \\
% \hline
% Power & 10W & 250W\\
% \hline
% \end{tabular}
% % \vspace{-0.2cm}
% \end{table}

%%%%%%%%%%%%%%%%%%%%%%%%%%%%%%%%%%%%%%%%%%%%%%%%%%%%%%%%%%%%%%%%%%%%%%%%%%%%%%%%%%%%%%%%%%%%%%%%%%%%%%%%%%%%%%%%%%%%%%%%%%%%%%%%%%%%%%%%%%%%%%%%%%%%%%%%%%%%%%%%%%%%%%%%
\vspace{-0.1cm}
\section{Results and Discussions}
\label{Sec_Results}
\vspace{-0.1cm}

%%%%%%%%%%%%%%%%
\subsection{Impact of Different Quantization Approaches on Accuracy}
\label{Sec_Results_Accuracy}
\vspace{-0.1cm}

\begin{figure*}[t]
% \vspace{-0.2cm}
\centering
\includegraphics[width=0.9\linewidth]{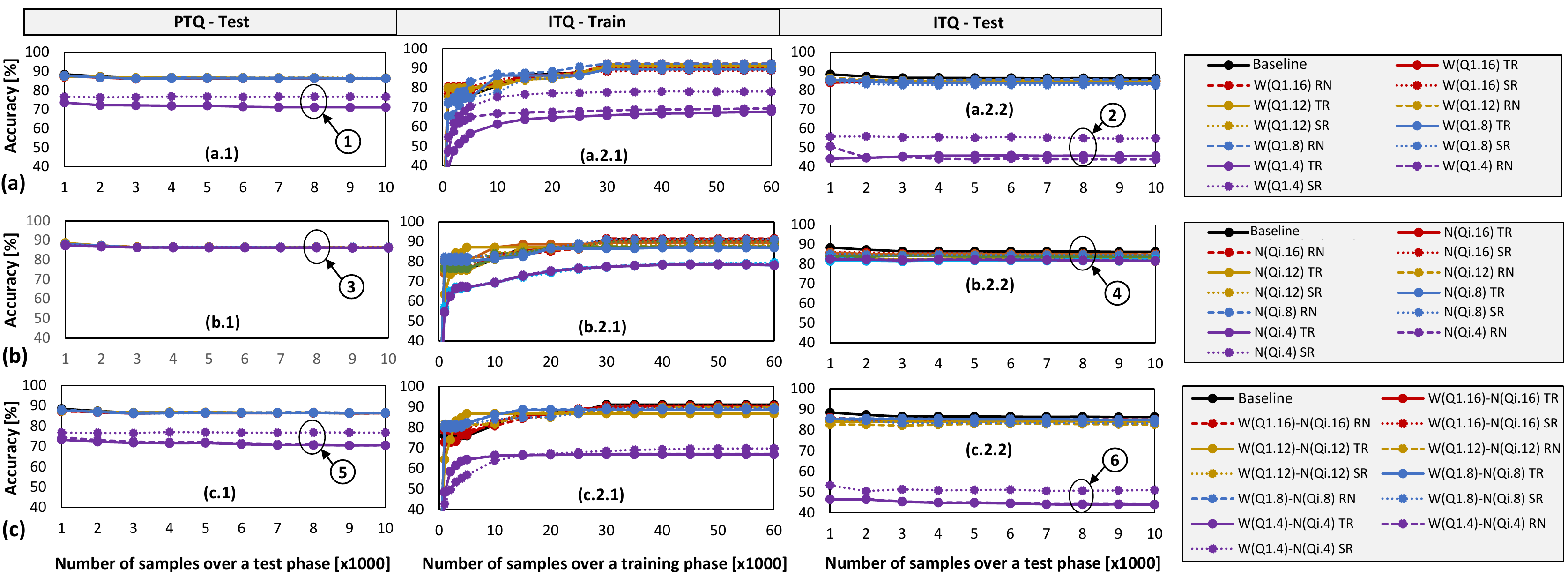}
\vspace{-0.3cm}
\caption{Results for the U-SNN with the MNIST dataset, when varying (a) the precision of weights and the rounding schemes, (b) the precision of neuron parameters and the rounding schemes, (c) the precision of weights and neuron parameters, as well as the rounding schemes, in terms of (a.1/b.1/c.1) the test accuracy in the PTQ, (a.2.1/b.2.1/c.2.1) the estimated accuracy during the training in the ITQ, and (a.2.2/b.2.2/c.2.2) the test accuracy in the ITQ.}
\label{Fig_Results_USNN_AccSamples}
\vspace{-0.3cm}
\end{figure*}

\textbf{Accuracy of the Unsupervised SNN:}
In the U-SNN case, we quantize the weights ($w$) and the neuron parameters (i.e., $V_{reset}$, $\theta$, $V$, and $V_{th}$), and the experimental results are shown in Fig.~\ref{Fig_Results_USNN_AccSamples}. 
Here, N(\texttt{Qi.f}) represents the precision of variables $V$ and $V_{th}$.
Notable accuracy degradation from the baseline accuracy is observed when the weights' bitwidth is reduced to the 4-bit fraction, as pointed by label-\circled{1} for the PTQ and label-\circled{2} for the ITQ.
The reason is that the 4-bit fraction (or fewer) for weights does not have sufficient levels of value to modulate the input spikes, thereby making the learning process ineffective. 
Meanwhile, quantizing $V$ and $V_{th}$ with the same fractional bits (i.e., 4 bits) still maintains the accuracy compared to the baseline, as shown by label-\circled{3} and label-\circled{4} for the PTQ and the ITQ, respectively. 
The reason is that the values for updating the $V$ and $V_{th}$ can be represented using fewer fractional bits than the ones for updating the weights $w$. 
These also indicate that the weights are more significant than the neuron parameters, as their small update can change the accuracy significantly.
Hence, quantizing all parameters of the U-SNN also leads to a notable accuracy degradation when the fractional bitwidth is reduced to 4 bits (or fewer), as pointed by label-\circled{5} and label-\circled{6} for the PTQ and the ITQ, respectively

%%%%
% \smallskip
% \subsubsection{Accuracy of the Supervised SNN}
% \label{Sec_Results_Accuracy_SSNN}

\textbf{Accuracy of the Supervised SNN:}
In the S-SNN case, we quantize the weights ($w$), and the neuron parameters (i.e., $\alpha$, $\beta$, $\gamma$, $P$, $Q$, $R$, and $V$), and the experimental results are presented in Fig.~\ref{Fig_Results_USNN_AccSamples}.  
Here, N(\texttt{Qi.f}) represents the precision of variables $P$, $Q$, $R$, and $V$.
Notable accuracy degradation from the baseline accuracy is observed when reducing the fractional bits of either the weights and the neuron parameters to 10 bits (and fewer), indicating that the weights and the neuron parameters have comparable significance to the accuracy. 
These also indicate that the S-SNN requires considerable bitwidth to maintain the high accuracy for the DVS-Gesture dataset.  
The reason is that, the DVS-Gesture is a relatively complex dataset because, besides considering the stream of events in each frame, the network has to draw a correct conclusion of a gesture for the complete stream of events from multiple frames.
Therefore, it requires considerable bitwidth to distinguish a gesture from other gestures in each frame and in a complete stream of events.  

\begin{figure*}[t]
% \vspace{-0.2cm}
\centering
\includegraphics[width=0.9\linewidth]{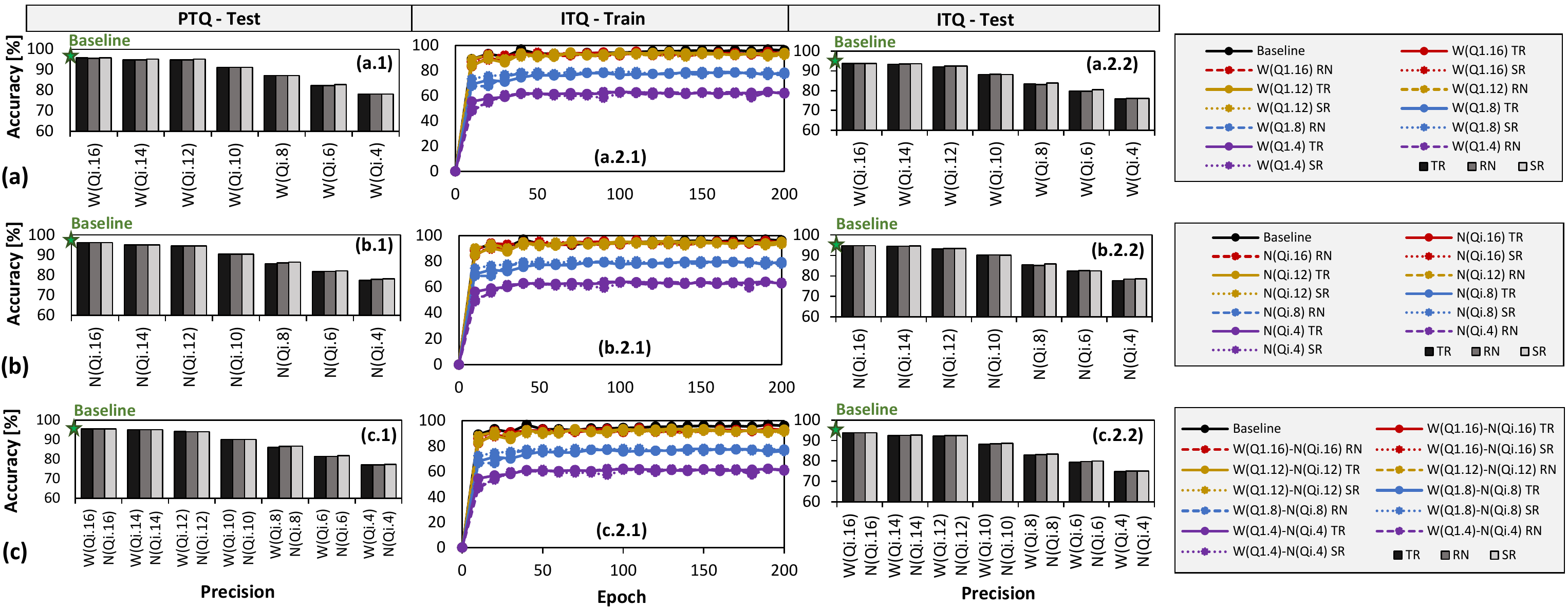}
\vspace{-0.4cm}
\caption{Results for the S-SNN with the DVS-Gesture dataset, when varying (a) the precision of weights and the rounding schemes, (b) the precision of neuron parameters and the rounding schemes, (c) the precision of weights and neuron parameters, as well as the rounding schemes, in terms of (a.1/b.1/c.1) the test accuracy in the PTQ, (a.2.1/b.2.1/c.2.1) the estimated accuracy during the training in the ITQ, and (a.2.2/b.2.2/c.2.2) the test accuracy in the ITQ.}
\label{Fig_Results_SSNN_AccSamples}
\vspace{-0.6cm}
\end{figure*}

% \smallskip
\textbf{Additional Discussion:} 
We also make the following observations across different network types (i.e., U-SNN and S-SNN) and different quantization approaches.
\begin{itemize}[leftmargin=*]  
    \item The SR scheme generally achieves slightly better accuracy than other rounding schemes, because this scheme is not biased towards a specific rounding direction, thereby having a higher probability of values that obtain higher accuracy. 
    However, it consumes the highest hardware resource as it needs a random number generator.  
    \item Different combinations of the quantization and rounding schemes achieve various accuracy, but their values are not significantly different. Users can decide the quantization and rounding schemes, as well as the parameters to be quantized, that are suitable for the target applications, 
    considering the accuracy and memory constraints, and the exploration cost.
    Therefore, the overhead depends on the selected scheme.
    % \item Different combinations of quantization schemes (PTQ and ITQ) and rounding schemes (TR, RN, and SR), achieve various accuracy, but their values are not significantly different. Therefore, \add{users can decide the quantization and rounding schemes, as well as the parameters to be quantized, that are suitable for the target applications, 
    % after the exploration process, 
    % considering the accuracy and memory constraints, and the exploration cost}.
\end{itemize}
%
% The above discussion shows that our framework provides a comprehensive information to the users regarding the accuracy of the given SNN models under different quantization schemes, different precision levels, and different rounding schemes.  
% Note, our focus is to evaluate the applicability of our Q-SpiNN on different SNN types (i.e., different architecture and learning approaches), hence we do not compare the performance (i.e., accuracy) between the unsupervised and supervised learning.

%%%%%%%%%%%%%%%%
\vspace{-0.2cm}
\subsection{SNN Model Selection with the Memory-Accuracy Trade-Off}
\label{Sec_Results_Select}
\vspace{-0.1cm}

To find the SNN model that offers a good memory-accuracy trade-off, we employ the proposed reward function in Eq.~\ref{Eq_Reward} that quantifies the trade-off benefit of the given model. 
To do this, we need to define the coefficient $\mu$ in the reward function. 
Small $\mu$ means that the function prioritizes the weight of accuracy more than the memory. 
On the other hand, large $\mu$ means that the function prioritizes the weight of memory more than the accuracy.
The users can define the value of $\mu$ based on their preferences to meet the design specifications.
In this work, for the exploration purpose, we define the value of coefficient $\mu \in \{0.01,0.1,0.2,0.3,0.4,0.5,1\}$.

\begin{figure}[hbtp]
\vspace{-0.3cm}
\centering
\includegraphics[width=0.9\linewidth]{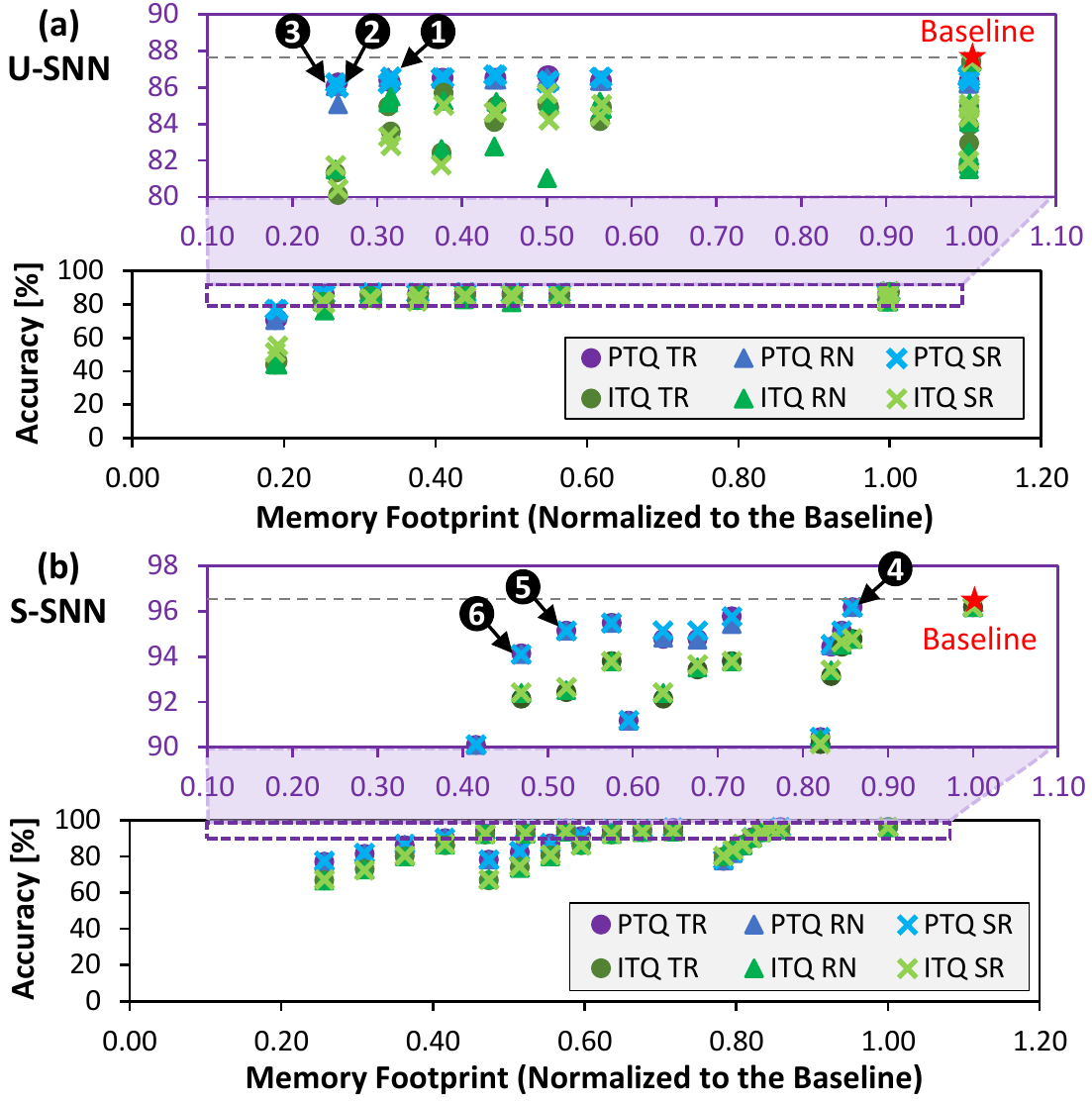}
\vspace{-0.3cm}
\caption{The accuracy vs. the normalized memory footprint for (a) the U-SNN and (b) the S-SNN.}
\label{Fig_Results_AccMem_SNNs}
\vspace{-0.5cm}
\end{figure}

\textbf{Model Selection for the Unsupervised SNN:}
We apply the proposed reward function to the explored U-SNN models and the results are provided in Fig.~\ref{Fig_Results_AccMem_SNNs}(a) and Table~\ref{Table_DSE}, from which we make the following observations.
\begin{itemize}[leftmargin=*]
    \item Label-\circledB{1}: The model that has the highest reward for $\mu = 0.01$ is the one that employs W(\texttt{Q1.8})-N(\texttt{Qi.8}) precision, and achieves 86.56\% accuracy and 3.194x memory saving. 
    \item Label-\circledB{2}: The model that has the highest reward for $\mu = 0.3$ is the one that employs W(\texttt{Q1.6})-N(\texttt{FP32}) precision, and achieves 86.27\% accuracy and 3.94x memory saving. 
    \item Label-\circledB{3}: The model that has the highest reward for $\mu \in \{0.1,0.2,0.4,0.5,1\}$ is the one that employs W(\texttt{Q1.6})- N(\texttt{Qi.6}) precision, and achieves 86.24\% accuracy and 3.987x memory saving. 
\end{itemize}
These results show that larger $\mu$ shifts the preferred model towards the one with smaller memory, which typically has lower accuracy. 
Meanwhile, smaller $\mu$ shifts the preferred model towards the one with higher accuracy, which typically has larger memory footprint.
If the maximum tolerance of accuracy degradation is 1\% from the baseline, then the model with W(\texttt{Q1.6})-N(\texttt{Qi.6}) precision is the Pareto-optimal one, with 86.24\% accuracy and 3.987x memory saving. 

% \smallskip
%%%%
% \subsubsection{Model Selection for the Supervised SNN}
% \label{Sec_Results_Select_SSNN}

\textbf{Model Selection for the Supervised SNN:}
We also apply the proposed reward function to the explored S-SNNs models and the results are provided in Fig.~\ref{Fig_Results_AccMem_SNNs}(b) and Table~\ref{Table_DSE}, from which we make the following observations.
\begin{itemize}[leftmargin=*]
    \item Label-\circledB{4}: The model that has the highest reward for $\mu = 0.01$ is the one that employs W(\texttt{FP32})-N(\texttt{Qi.16}) precision, and achieves 96.14\% accuracy and 1.165x memory saving. 
    \item Label-\circledB{5}: The model that has the highest reward for $\mu = 0.1$ is the one that employs W(\texttt{Q1.14})-N(\texttt{Qi.14}) precision, and achieves 95.14\% accuracy and 1.926x memory saving. 
    \item Label-\circledB{6}: The model that has the highest reward for $\mu \in \{0.2,0.3,0.4,0.5,1\}$ is the one that employs W(\texttt{Q1.12})- N(\texttt{Qi.12}) precision, and achieves 94.14\% accuracy and 2.132x memory saving. 
\end{itemize}
Here, similar trend regarding the impact of $\mu$ is also observed, e.g., larger $\mu$ shifts the preferred model towards the one with smaller memory and lower accuracy. 
If the maximum tolerance of accuracy degradation is only 1\% from the baseline, then the model with W(\texttt{FP32})-N(\texttt{Qi.16}) precision is selected.
% (with 96.14\% accuracy and 1.926x memory saving)
If we relax the tolerance to 2\%, it suggests different Pareto-optimal SNN model, i.e., the model with W(\texttt{Q1.14})-N(\texttt{Qi.14}) precision, 95.14\% accuracy, and 1.926x memory saving. 
% If we relax the tolerance even more, e.g., allowing 3\% accuracy degradation, then the one with W(\texttt{Q1.12})-N(\texttt{Qi.12}) precision (with 94.14\% accuracy and 2.132x memory saving) is selected. 

The above results and discussion show that our Q-SpiNN framework provides (1) a comprehensive information about the accuracy and the memory of the given SNN models under different quantization approaches, and (2) an effective model selection to find the efficient SNN model. 
Moreover, the users can set $\mu$ with their preferred value in the reward function to select an SNN model that meets their design requirements.

% \smallskip

%%%%%%%%%%%%%%%%%%%%%%%%%%%%%%%%%%%%%%%%%%%%%%%%%%%%%%%%%%%%%%%%%%%%%%%%%%%%%%%%%%%%%%%%%%%%%%%%%%%%%%%%%%%%%%%%%%%%%%%%%%%%%%%%%%%%%%%%%%%%%%%%%%%%%%%%%%%%%%%%%%%%%%%%
\section{Conclusion}
\label{Sec_Conclusion}
\vspace{-0.1cm}

We propose the Q-SpiNN framework for quantizing SNNs through (1) quantization of different parameters, (2) exploration that considers different quantization schemes, precision levels, and rounding schemes, and (3) employment of a reward function for model selection. 
For the unsupervised SNN, the Q-SpiNN obtains ca. 4x memory saving and keeps the accuracy within 1\% from the baseline on the MNIST. 
For the supervised one, it obtains ca. 2x memory saving, and keeps the accuracy within 2\% from the baseline on the DVS-Gesture.
Therefore, our framework would enable the SNN systems to be deployable on the resource-constrained devices.

\begin{table}[hbtp]
% \vspace{-0.4cm}
\caption{The best accuracy and the memory savings, across different quantization schemes, precision levels, and rounding schemes}
\label{Table_DSE}
\vspace{-0.2cm}
\centering
\scriptsize
% \footnotesize
\begin{tabular}{|c|c|c|c|c|}
\hline
\multirow{3}{*}{\textbf{Precision}}&\multicolumn{2}{c|}{\textbf{U-SNN}} & \multicolumn{2}{c|}{\textbf{S-SNN}} \\
\cline{2-5} 
 & \multirow{2}{*}{\textbf{Accuracy}} & \textbf{Memory} & \multirow{2}{*}{\textbf{Accuracy}} & \textbf{Memory} \\ 
 &  & \textbf{Saving} &  & \textbf{Saving} \\
\hline
\hline
Baseline & 87.20\% & 1.000x & 96.18\% & 1.000x \\
W(Q1.16) & 86.54\% & 1.771x & 95.79\% & 1.395x \\
W(Q1.14) & 86.67\% & 1.990x & 95.14\% & 1.479x \\
W(Q1.12) & 86.66\% & 2.271x & 95.14\% & 1.573x \\
W(Q1.10) & 86.52\% & 2.644x & 91.18\% & 1.680x \\
W(Q1.8) & 86.56\% & 3.165x & 87.14\% & 1.803x \\
W(Q1.6) & \textbf{86.27\%} & 3.940x & 82.75\% & 1.934x \\
W(Q1.4) & 76.66\% & 5.218x & 78.14\% & 2.110x \\
N(Qi.16) & 86.36\% & 1.002x & \textbf{96.14\%} & \textbf{1.165x} \\
N(Qi.14) & 86.63\% & 1.002x & 95.14\% & 1.182x \\
N(Qi.12) & 86.53\% & 1.002x & 94.53\% & 1.200x \\
N(Qi.10) & 86.53\% & 1.003x & 90.44\% & 1.218x \\
N(Qi.8) & 86.67\% & 1.003x & 86.44\% & 1.238x \\
N(Qi.6) & 86.57\% & 1.003x & 82.10\% & 1.257x \\
N(Qi.4) & 86.58\% & 1.003x & 78.10\% & 1.277x \\
W(Q1.16)-N(Qi.16) & 86.53\% & 1.778x & 95.49\% & 1.739x \\
W(Q1.14)-N(Qi.14) & 86.67\% & 1.999x & \textbf{95.14\%} & \textbf{1.926x} \\
W(Q1.12)-N(Qi.12) & 86.68\% & 2.284x & \textbf{94.14\%} & \textbf{2.132x} \\
W(Q1.10)-N(Qi.10) & 86.52\% & 2.663x & 90.10\% & 2.404x \\
W(Q1.8)-N(Qi.8) & \textbf{86.56\%} & \textbf{3.194x} & 86.79\% & 2.756x \\
W(Q1.6)-N(Qi.6) & \textbf{86.24\%} & \textbf{3.987x} & 81.79\% & 3.228x \\
W(Q1.4)-N(Qi.4) & 76.71\% & 5.306x & 77.40\% & 3.895x \\
\hline
\end{tabular}
\vspace{0.18cm}
\end{table}


% \vspace{-0.1cm}
\section*{Acknowledgment}
\vspace{-0.1cm}
This work was partly supported by Indonesia Endowment Fund for Education (LPDP) Scholarship Program, from the Ministry of Finance, Indonesia. 
% \vspace{-0.1cm}

\bibliographystyle{IEEEtran}
\bibliography{bibliography}

\end{document}